\documentclass[sigconf,anonymous=false]{acmart}
\pdfoutput=1

\setcopyright{none}
\renewcommand\footnotetextcopyrightpermission[1]{}
\usepackage{multirow}
\usepackage{url}
\usepackage{graphicx}
\usepackage{amsmath}
\usepackage{amsfonts}

\usepackage{algorithm}
\usepackage{lipsum}  
\usepackage{algorithmic}
\usepackage{booktabs}
\usepackage{subcaption}
\usepackage{booktabs}

\DeclareMathOperator*{\argmin}{\arg\!\min}
    
\begin{document}

\title{On genetic programming representations and fitness functions for interpretable dimensionality reduction}



\author{Thomas Uriot}
\affiliation{%
  \institution{Leiden University Medical Center}
  \city{Leiden} 
  \country{The Netherlands} 
}

\author{Marco Virgolin}
\affiliation{%
  \institution{Centrum Wiskunde \& Informatica}
  \city{Amsterdam} 
  \country{The Netherlands} 
}

\author{Tanja Alderliesten}
\affiliation{%
  \institution{Leiden University Medical Center}
  \city{Leiden} 
  \country{The Netherlands} 
}

\author{Peter A.N. Bosman}
\affiliation{%
  \institution{Centrum Wiskunde \& Informatica}
  \city{Amsterdam} 
  \country{The Netherlands} 
}

\begin{abstract}
Dimensionality reduction (DR) is an important technique for data exploration and knowledge discovery. However, most of the main DR methods are either linear (e.g., PCA), do not provide an explicit mapping between the original data and its lower-dimensional representation (e.g., MDS, t-SNE, isomap), or produce mappings that cannot be easily interpreted (e.g., kernel PCA, neural-based autoencoder). Recently, genetic programming (GP) has been used to evolve interpretable DR mappings in the form of symbolic expressions. There exists a number of ways in which GP can be used to this end and no study exists that performs a comparison.
In this paper, we fill this gap by comparing existing GP methods
as well as devising new ones.
We evaluate our methods on several benchmark datasets based on predictive accuracy and on how well the original features can be reconstructed using the lower-dimensional representation only. Finally, we qualitatively assess the resulting expressions and their complexity. 
We find that various GP methods can be competitive with state-of-the-art DR algorithms and that they have the potential to produce interpretable DR mappings.
\end{abstract}



\maketitle
\pagestyle{plain}

\section{Introduction}

%


Dimensionality reduction (DR) is a key instrument for knowledge discovery~\cite{huang2009dimensionality,katole2015new}. 
Informally, we can define DR (see Section \ref{sec:dimensionality_reduction} for a formal definition) to be the area of study concerned with finding lower dimensional representations of the original data, such that some meaningful properties 
of the original dataset are preserved. 
It is based on the fact that most real-world datasets only have artificially high dimensionality due to, for instance, interdependent or noisy features. However, current state-of-the-art DR algorithms act as a black-box when going from high to low dimensions (e.g., t-SNE~\cite{maaten2008visualizing}, isomap~\cite{roweis2000nonlinear}, LLE~\cite{roweis2000nonlinear}), and do not provide any means to inspect the resulting lower dimensions in terms of the original variables, i.e., they do not provide a functional mapping that describes how the original dimensions are compressed into the lower-dimensional (or also called latent) representation. 
Other methods, such as kernel PCA~\cite{scholkopf1998nonlinear} and neural autoencoders, do provide a functional mapping between the original data and its lower dimensional representation, however these mappings are hardly interpretable.
Furthermore, in the absence of meaningful labels or classes, it becomes even more challenging to assess and inspect whether a lower-dimensional representation produced by DR is truly meaningful. 
Finally, DR is often used for data exploration, in which case the ability to draw insights from the data is key. 
For these reasons, producing a functional mapping that is human-interpretable can be very important, as that allows to assess whether the compression the mapping performs may be reasonable and, e.g., safe to use in high-stakes applications of machine learning~\cite{adadi2018peeking}.


Recently, the use of genetic programming~\cite{koza1992genetic} (GP) has been an increasingly popular way to produce interpretable models. For instance, GP has been used to learn programs encoded by simple symbolic expression representing physical laws~\cite{cranmer2020discovering,izzo2017differentiable}. Furthermore, recent works have investigated the use of such symbolic expressions produced by GP as interpretable models~\cite{evans2019s,lensen2019can,virgolin2020explaining,virgolin2020learning}. In this work, we leverage the interpretability potential of GP and apply it to non-linear DR. 

GP has been shown to be a promising avenue to produce interpretable lower-dimensional representations in several recent works~\cite{lensen2019can,lensen2020genetic,lensen2020multi}. However, more representations and evaluation schemes can be imagined, but no work has been done to provide a comparison of these different options. 
We attempt to fill this gap by designing new GP-based methods for DR and establish a robust evaluation protocol across two objectives (predictive power of the newly constructed features and their ability to reconstruct the original input). The main contributions of this paper are threefold: (i) we investigate how GP can best be used for DR by considering standard DR loss functions and GP representations (see Section \ref{sec:method}), (ii) additionally, we perform experiments with several DR loss functions that have previously never been applied to GP, such as weighted rank-correlation, using a neural-based autoencoder teacher, or directly using a GP-based autoencoder to reconstruct the input, and (iii), we show that our GP-based methods are able to learn non-linear and concise formulas of the lower-dimensional latent space.
We benchmark our methods against three mainstream DR algorithms: principal components analysis~\cite{jolliffe2016principal} (PCA), locally linear embedding~\cite{roweis2000nonlinear} (LLE), and isomap~\cite{tenenbaum2000global}. 

\section{Background}
\label{sec:background}

\subsection{Dimensionality Reduction}
\label{sec:dimensionality_reduction}

The goal of DR is to find a lower-dimensional representation of the original data such that some properties of the original data, such as total variance (e.g., PCA), distances between points (e.g., isomap), ability to reconstruct the original features (e.g., autoencoders) are preserved. Formally, let us denote the original dataset by $\mathcal{X} \in \mathbb{R}^{n \times p}$ where $n$ and $p$ represent the number of instances and features, respectively. Then, the goal is to find another representation $\widetilde{\mathcal{X}} \in \mathbb{R}^{n \times k}$ of $\mathcal{X}$, such that $k \ll p$, and such that the new representation retains some meaningful properties of the original data. Note that throughout this paper, we will refer to $X^j$ and $\widetilde{X}^j$ as the $j^{\textrm{th}}$ dimension (feature), and $x_i$ and $\Tilde{x}_i$ as the $i^{\textrm{th}}$ observation of the original and lower-dimensional representation, respectively.

\subsection{Genetic Programming for Dimensionality Reduction}

Recently, DR using GP has gained traction and has started being more thoroughly explored. In~\cite{lensen2019can}, the authors use GP to perform DR by using a fitness function that encourages the preservation of local neighborhoods. In~\cite{lensen2020multi}, the authors build on their previous work~\cite{lensen2019can} by taking a bi-objective approach where the first objective encourages to preserve local structure, while the second objective represents the number of dimensions in the latent representation. However, both in~\cite{lensen2019can} and~\cite{lensen2020multi}, there is no constraint imposed on the complexity of the evolved expressions, potentially rendering them non-interpretable. This issue is addressed in~\cite{lensen2020genetic} where the authors use a bi-objective approach with the first objective being the cost function used in the t-SNE algorithm~\cite{maaten2008visualizing} and the second objective being a proxy for complexity. However, t-SNE is devised to strictly preserve local neighborhoods and does not result in a functional mapping, thus it is unclear whether the cost function of t-SNE is well-suited to be used for GP. 

In addition, GP has recently been used to perform feature extraction and transfer learning by constructing non-linear combinations of features to then be used in subsequent supervised learning tasks~\cite{lensen2017improving,virgolin2020explaining,munoz2020transfer}. This type of application requires to specify a predictive task with ground truth labels. However, we may be interested in discovering intrinsic relationships between the variables or find an interpretable functional mapping for the compression of the original dataset without necessarily having an associated predictive task. One way to tackle the aforementioned problem was explored in~\cite{schofield2020evolving}, where the authors used GP to predict the value of each feature by using the remaining ones as input. However, this process does not belong to the realm of DR since the number of features remains the same. 

\section{Methodology}
\label{sec:method}

In GP, the main way to represent symbolic expressions is by defining it as a computational tree~\cite{poli2008field}. In this paper, we follow the literature and use trees to represent our GP programs. A detailed description of the tree-based representations we consider is given in the following subsection.

\begin{figure*}[htb]
\centering
    \includegraphics[width=1.0\linewidth]{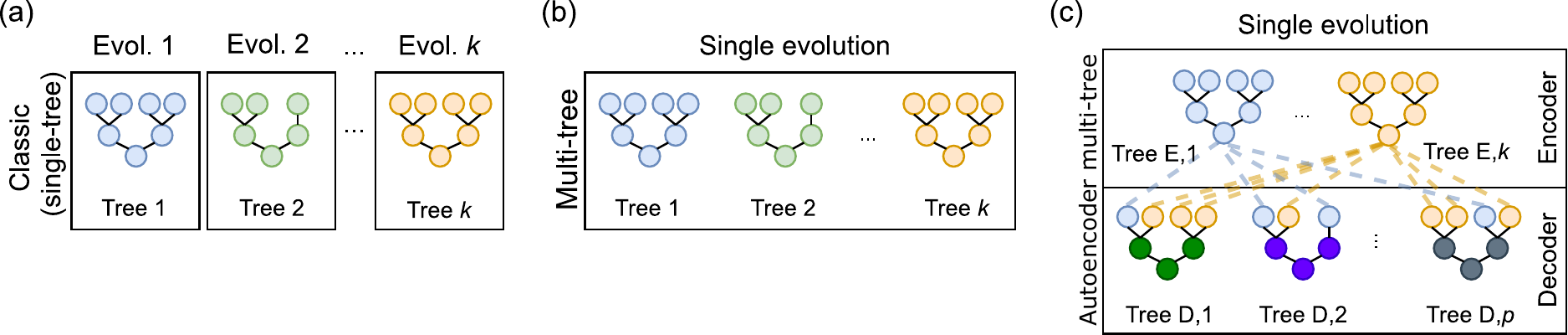}
    \caption{Three ways of mapping the original data to its lower-dimensional representation using computational trees. Note that the trees are read top-down, from leaves to output. \textbf{(a) Classic:} A GP individual is represented by a single tree. Each tree maps the input to a unique latent dimension, and is evolved separately. It thus requires $k$ independent runs.
    \textbf{(b) Multi-tree:} A GP individual is a collection of $k$ trees that are evolved jointly in a single evolution run.
    \textbf{(c) Autoencoder multi-tree multi-tree:} A GP individual is made of two parts: the encoder and the decoder. The encoder is a multi-tree composed of $k$ trees and is responsible to map the original input to its lower-dimensional representation. The decoder is a multi-tree composed of $p$ trees, where $p$ is the number of original features, and is responsible to map the output of the encoder to reconstruct the original features. The encoder and the decoder evolve jointly.
    }
    \label{fig:tree_representation}
\end{figure*}



\subsection{Computational Tree Representations}
\label{sec:multi_tree_representation}

\subsubsection{Single tree}

The single-tree representation, as shown in Figure \ref{fig:tree_representation}a, is one of the simplest way to represent a symbolic expression. In the single-tree representation, a single tree is responsible to output a unique single value. Formally, if we denote a single tree by $m$, we have that $m_j: \mathbb{R}^{p} \to \mathbb{R}$, for $j=1,\ldots,k$. In words, each tree maps the original dimensions (the terminals at the leaves) to a single latent dimension (the root of the tree). 

This representation is limited when it comes to modeling multi-output functions which is the case in DR, if $k>1$. Indeed, each latent dimension is evolved separately (see Figure \ref{fig:tree_representation}a). Therefore, this representation cannot efficiently model non-separable fitness functions, where variables are dependent and should thus be jointly optimized, as is the case in DR.

\subsubsection{Classic multi-tree}

Similarly to~\cite{lensen2019can,lensen2020genetic,lensen2020multi}, we use a multi-tree representation as illustrated in Figure \ref{fig:tree_representation}b. Each tree of the multi-tree takes the original data as input and outputs a single dimension of the lower-dimensional representation. During evolution, the trees of the multi-tree evolve simultaneously, different from single-tree representation, where each tree evolves separately during independent runs. This representation has two main advantages: (i) it only requires a single evolution run, (ii) it is able to learn non-separable fitness functions, where the variables are dependent, which is the case in cost functions that are typically used in DR. 

In this paper, we will denote the multi-tree representation by $\mathcal{MT(\cdot)}=\{m_j(\cdot)\}^{k}_{j=1}$, where $m_j$ represents the $j^{\textrm{th}}$ tree of the multi-tree and $k$ is the number of dimensions in the lower-dimensional representation.


\subsubsection{Autoencoder multi-tree} 
\label{sec:gp_autoencoder}

In addition to the classic multi-tree, we propose a new representation for DR (inspired from automatically defined functions~\cite{koza1994genetic}), composed of two classic multi-trees, with one multi-tree acting as encoder and one as decoder. Specifically, the encoder is a multi-tree composed of $k$ trees and is responsible to map the original input to its lower-dimensional representation. On the other hand, the decoder is a multi-tree composed of $p$ trees where $p$ is the number of original features. It is responsible to map the output of the encoder to reconstruct the original features. The encoder and the decoder are jointly evolved in a single run. Formally, let us denote the encoder and decoder of the autoencoder multi-tree by $\mathcal{AMT}^{(e)}$ and $\mathcal{AMT}^{(d)}$ respectively, then we have that $\mathcal{AMT}^{(e)}(\cdot) = \{m^{{(e)}}_j(\cdot)\}^{k}_{j=1}$ and  $\mathcal{AMT}^{(d)}(\cdot) = \{m^{{(d)}}_j(\cdot)\}^{p}_{j=1}$, where  $m^{{(e)}}_j: \mathbb{R}^{p} \to \mathbb{R}$ and  $m^{{(d)}}_j: \mathbb{R}^{k} \to \mathbb{R}$.

The purpose of the autoencoder multi-tree is to test whether a fully GP-based approach, i.e., one where both the encoder and decoder are evolved at the same time, is capable of performing DR effectively, as explained in the following subsection. 

\subsection{Fitness Functions}
\label{sec:cost_functions}

We investigate several fitness functions, denoted by $\mathcal{F}$, to guide the evolutionary process of the multi-tree GP. As mentioned in Section \ref{sec:dimensionality_reduction}, we denote the lower-dimensional representation of the original dataset $\mathcal{X}$ by $\mathcal{\widetilde{X}}$, where the $i^{\textrm{th}}$ data point is given by $\widetilde{x}_{i} = \mathcal{MT}(x_i)$, $i=1,\ldots,n$, for a given multi-tree individual $\mathcal{MT}$. Furthermore, we denote the distance matrices of the original dataset and its lower-dimensional representation by $D$ and $\widetilde{D}$, respectively, where $D_{i,j}=d_{i,j}$ and $\widetilde{D}_{i,j}=\widetilde{d}_{i,j}$. For the distance preserving and rank preserving fitness functions, we investigate two distance metrics: (i) Euclidean distance, and, (ii) geodesic distance. The geodesic distance is particularly suited to represent the local low-dimensional geometry of the data manifold~\cite{tenenbaum2000global} that the Euclidean distance may not be able to capture.

Formally, for a fitness function denoted by $\mathcal{F(\mathcal{X}, \cdot)}$, we seek $\mathcal{MT}^{*}$ such that: 

\begin{equation}
\mathcal{MT}^{*} = \argmin_{\mathcal{MT}} \mathcal{F(\mathcal{X}, \cdot)}.
\end{equation}

Note that for the GP-based autoencoder fitness function, we use an autoencoder multi-tree and the best individual would thus be denoted by $\mathcal{AMT}^*$. 

\subsubsection{Distance preserving}
\label{sec:dist_preserving}

This fitness function is based on preserving distances between pair of instances when mapping the original data to its lower-dimensional representation. To this end, we use Sammon mapping~\cite{sammon1969nonlinear} which aims to minimize the following:

\begin{equation}
   \mathcal{F}_{\textrm{dist}} = \mathcal{F(\mathcal{X}, \mathcal{\widetilde{X}})} = \frac{1}{\sum_{i<j}d_{i,j}}\sum_{i<j}\frac{(d_{i,j}-\widetilde{d}_{i,j})^2}{d_{i,j}}.
\end{equation}

The effect of $d_{i,j}$ in the denominator is to put emphasis on points that are close to each-other in the original space. In other words, we give more importance in preserving local structure than global.

\subsubsection{Rank preserving}
\label{sec:rank_cost_function}

This fitness function is similar to preserving distances except we are only interested in the ranks of the distances. Essentially, this is a measure of the monotonicity of the relationship between the distances in the original and lower-dimensional representations. This has the advantage of being more flexible, less sensitive to outliers (i.e., large distances) and thus more easily learnable than the distance preserving fitness. In~\cite{lensen2020multi}, the authors use Spearman's rank correlation coefficient~\cite{zwillinger1999crc} as the fitness function to perform dimensionality reduction using GP. In this paper, we use Kendall's $\tau$ which is preferred to Spearman's rank correlation coefficient for smaller samples~\cite{kendall1938new}. Our aim is to maximize Kendall's $\tau$: 

\begin{equation}
    \tau = \frac{n_c-n_d}{\frac{N}{2}(N-1)},
\end{equation}

where $n_c$ is the number of \emph{concordant pairs}, $n_d$ is the number of \emph{discordant pairs}, and $N$ is the total number of pairs. Let us denote the $i^{\textrm{th}}$ row of the distance matrices $\widetilde{D}$ and $D$ by $\Tilde{d}_i$ and $d_i$ respectively. Then, we have that a pair is concordant if $\textrm{sgn}(d_{i,j}-d_{i,l})=\textrm{sgn}(\widetilde{d}_{i,j}-\widetilde{d}_{i,l})$, and discordant if $\textrm{sgn}(d_{i,j}-d_{i,l})=-\textrm{sgn}(\widetilde{d}_{i,j}-\widetilde{d}_{i,l})$, for $j\neq l$ and $\textrm{sgn}(\cdot)$ being the standard signum function.

Taking the average across instances, the final fitness function (to be minimized) is defined as:

\begin{equation}
   \mathcal{F}_{\textrm{rank}} = \mathcal{F(\mathcal{X}, \mathcal{\widetilde{X}})} = -\frac{1}{n}\sum_{i=1}^n\tau_i = -\frac{1}{n}\sum_{i=1}^n \frac{n_{c_{i}}-n_{d_{i}}}{\frac{N_i}{2}(N_i-1)},
\end{equation}

where $n_{c_{i}}$, $n_{d_{i}}$, and $N_i$ are computed using $d_i$ and $\widetilde{d}_i$.

However, this approach fails to take into account that discordances between observations with high rank (a rank of 0 being the highest rank) are more important than those between items with low rank. This is because we want to emphasize preserving local structure over global structure, and so, if the distance between two points in the original space is large (i.e. the rank is low), we want it to have a lesser importance in the final fitness function. This serves similar purpose to the denominator in the Sammon mapping in Equation (2). 

Therefore, we use a weighted version of Kendall's $\tau$ proposed in~\cite{vigna2015weighted}, where the weighting is defined by ranking the original distances and assigning the highest rank of 0 to the shortest distance, and so on. Then, a weighting function assigns a weight $w$ depending on the rank $r$. In this paper, we use a hyperbolic weighting function, and thus, we have that $w=\frac{1}{r+1}$.

\subsubsection{Neural-based autoencoder teacher}
\label{sec:neural_teacher}
This fitness function is inspired by the student-teacher literature, where a (usually) simple model (e.g., linear regression, low-depth trees, symbolic regression) is trained to approximate the output of another (complex) model. This is a two-step process where we first train a base model and then train the surrogate model on the output (or hidden layer) produced by the base model.


In our case, let us denote the latent layer of the neural-based autoencoder by $\mathcal{L}=\{l_j\}^{k}_{j=1}$, where $l_j$ represents the $j^{\textrm{th}}$ neuron and $k$ is the number of latent dimensions (e.g., the number of neurons in the latent layer). Our fitness function is then given by:

\begin{equation}
    \mathcal{F}_{\textrm{AE}} = \mathcal{F(\mathcal{X}, \mathcal{\widetilde{X}})} = \frac{1}{nk}\sum^{n}_{i=1}\sum^{k}_{j=1}(l_j(x_i)-\widetilde{x}_{i,j})^2,
\end{equation}

where $\widetilde{x}_{i,j} = m_j(x_i)$ is the output of the $j^{\textrm{th}}$ tree of the multi-tree $\mathcal{MT}=\{m_j\}^{k}_{j=1}$. In words, the fitness function is the mean-squared-error between the output of the multi-tree and the latent layer of the autoencoder.

\subsubsection{GP-based autoencoder}
\label{sec:gp_based_autoencoder}

This fitness function is similar to the neural-based autoencoder teacher fitness function in that it also makes use of an autoencoder. However, instead of evolving a classic multi-tree GP to match the latent layer of a neural autoencoder teacher, we let GP evolve a complete autoencoder, using the representation shown in Figure \ref{fig:tree_representation}c. As mentioned, the GP-based autoencoder is composed of two multi-trees representing the encoder and the decoder of the autoencoder. In this case, the lower-dimensional representation $\widetilde{X}$ of the original dataset $\mathcal{X}$ is given by the output of the encoding multi-tree. Formally, using the notation introduced in \ref{sec:gp_autoencoder}, we have that $\widetilde{x}_i = \mathcal{AMT}(x_i)^{(e)}$, $i=1,\ldots,n$, and the fitness function is given by:

\begin{equation}
    \mathcal{F}_{\textrm{GP}} = \mathcal{F(\mathcal{X}, \mathcal{\widehat{X}})} = \frac{1}{np}\sum^{n}_{i=1}\sum^{p}_{j=1}(x_i-\widehat{x}_{i,j})^2,
\end{equation}

where $\widehat{X}_{i,j} = \widehat{x}_{i,j} = m_j^{{(d)}}(x_i)$ is the output of the $j^{\textrm{th}}$ tree of the GP's decoder $\mathcal{AMT}^{(d)}$. In other words, the cost function is the reconstruction error between the output of the decoder and the original data, and the resulting lower-dimensional representation is given by the output of the encoder.

\subsection{Baseline Methods}
\label{sec:baselines}

We compare the performance of the GP-based methods with three dimensionality reduction baselines: principal components analysis~ (PCA), locally linear embedding (LLE), and isomap. These methods are chosen because they learn a mapping on the training data that can be applied to unseen data, which is not possible with other methods such as t-SNE or multi-dimensional scaling (MDS). This enables a fair comparison since our GP-methods are trained to perform dimensionality reduction, and the evaluation is done on unseen data, as depicted in Figure \ref{fig:evaluation_protocol}. In addition, the baselines are heterogeneous as PCA seeks a global orthogonal linear transformation of the data, while LLE is a local and non-linear method, and isomap seeks to preserve the estimated intrinsic geometry of the data manifold using geodesic distance.

\subsection{Performance Metrics}
\label{sec:eval_functions}

In this section, we discuss the choices made to evaluate the performance of the various fitness functions in guiding the evolution process of GP. This amounts to evaluating the quality of the resulting lower-dimensional representation of the original data. This is still an open problem in data science and will depend on the application and the questions that one wants to answer during the analysis. In addition, the fitness function being optimized during the dimensionality reduction process may be biased towards our evaluation metric or even be misaligned with it.

\subsubsection{Predictive performance}

In some situations, there exists a metric that truly represents our end goal, such as predictive performance on a supervised downstream task. If there exists such an objective (task) for a given dataset, then we do not have to worry whether the metric (e.g., classification accuracy or regression error) is biased towards any of our GP methods since it is a metric that we are \textit{objectively} interested in. Using the predictive power of the low-dimensional representation on downstream supervised tasks to evaluate dimensionality reduction algorithms has been done in recent studies~\cite{lensen2019can,lensen2020genetic,lensen2020multi}.

However, there are three main drawbacks with this approach. Firstly, we have to assume that the response (e.g., class structure for classification or target distribution for regression) is a significant factor within the manifold structure. Arguably, this assumption holds for the majority of curated datasets, and datasets for which supervised learning algorithms can perform well (i.e., one can learn labels from the data). Secondly, there may be some structure in the data that is not captured by the labels but that may still be of interest for knowledge discovery, and thus using predictive performance as a proxy would be insufficient. Finally, and perhaps the most critical drawback is that we may have datasets on which we want to perform dimensionality reduction but that have no labels.

In this work, we use balanced accuracy~\cite{brodersen2010balanced} as a performance metric. In particular, we follow~\cite{lensen2019can,lensen2020multi} and use a random forest classifier as evaluation model for accuracy (denoted by $E$ in Section \ref{sec:eval_protocol}), trained on the low-dimensional representation to predict the labels and compute the predictive performance.

\subsubsection{Reconstruction error}

Besides accuracy, we also consider the reconstruction error, as an additional metric, and, in this case, we use a neural-based decoder as the evaluation model. Note that, however, the reconstruction error is biased to favor the methods using the neural-based autoencoder teacher ($\mathcal{F}_{\textrm{AE}}$) and the GP-based autoencoder  ($\mathcal{F}_{\textrm{GP}}$) fitness functions since they both minimize the reconstruction error during optimization.

\section{Experimental setup}
\label{sec:experimental_setting}

\subsection{Datasets}
We test our methods on 5 datasets from the UCI repository~\cite{Dua:2019} for which a brief description is provided in Table 1. Note that for the purpose of our analysis, we only keep the continuous attributes, which is why the number of features is different from the original one for datasets that contain categorical variables. This is because we only use arithmetic operators and do not use any conditional or logical operators (e.g., \textit{if}, \textit{and}, \textit{or}) in our GP representation (see Section \ref{sec:hyperparameters}). In addition, some of our fitness functions are based on the Euclidean distance, which would not be appropriate for a mix of continuous and categorical variables.

\begin{table}[h!]
\centering
\caption{Datasets considered, where $n$, $d$, and $c$ represent the number of instances, features, and classes, respectively.}
\label{tab:datasets}

\begin{tabular}{lccc}
       \toprule  
        \textbf{Name}& $\mathbf{n}$ & $\mathbf{d}$ & $\mathbf{c}$ \\
            \midrule
            Ionosphere & 350 & 34 & 2  \\
            German Credit & 1000 & 24 & 2 \\
            Libras Movement & 359 & 90 & 15   \\
            Segmentation & 2310 & 19 & 7  \\
            Telescope (MGT) & 19019 & 10 & 2 \\
            \bottomrule
        \end{tabular}

\end{table}

\begin{figure*}[ht!]
    \includegraphics[width=0.85\linewidth]{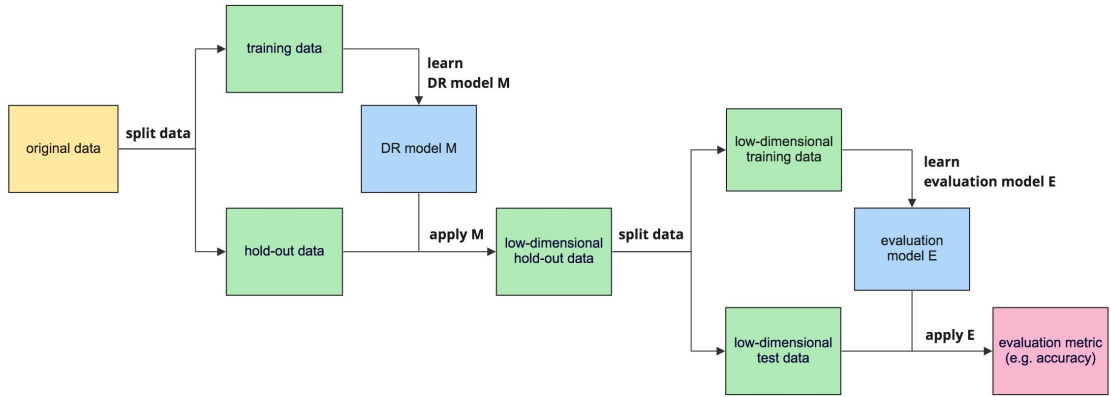}
    \caption{Evaluation protocol for our DR pipeline. First, we learn a DR model $M$ on a subset of the original data. This model is then applied on the held-out subset of the original data to obtain a lower-dimensional representation. We then learn an evaluation model $E$ on a subset of the lower-dimensional data and apply it to the held-out subset in order to compute the final performance metric.}
    \label{fig:evaluation_protocol}
\end{figure*}

\subsubsection{Principal component initialization}
\label{sec:pca_init}

Note that the fitness functions described in Section \ref{sec:cost_functions} make use of all the variables from the original data. This happens either at the time of computing the distance matrices or when computing the reconstruction error. In doing so, we may be using noisy variables as well as redundant variables (e.g., co-linear variables) which could be dominating other important variables during the optimization process. A simple and well established solution is to perform PCA before applying any sort of analysis that requires to compute distances between points, such as clustering analysis~\cite{ben2003detecting,ding2004k}. Thus, in this paper, we transform the original dataset using PCA (retaining $99\%$ of the original variance) to compute the fitness functions in Section \ref{sec:cost_functions}. This amounts to replacing $\mathcal{X}$ with $\mathcal{X}_{\textrm{pca}}$ in $\mathcal{F(\mathcal{X}, \cdot)}$, where $\mathcal{X}_{\textrm{pca}}$ is the transformation of $\mathcal{X}$ after applying PCA and keeping a sufficient number of principal components to account for $99\%$ of the original variance. Note that in order to keep the GP expressions interpretable in terms of the original input, the input to the multi-tree GP is still the original data $\mathcal{X}$, as opposed to the PCA transformed data $\mathcal{X}_{\textrm{pca}}$.

\subsection{Genetic Programming Parameters}
\label{sec:hyperparameters}

We use standard GP parameter settings, as shown in Table \ref{tab:gp_parameters}. A basic tournament selection~\cite{poli2008field} process is used to chose the individuals that will act as parents to produce offspring via the variation operators. Each parent can be subject to up to three different genetic operators: (i) \textit{crossover}, (ii) \textit{subtree mutation}, and (iii) \textit{one-point mutation}, with probabilities of $p_c$, $p_s$, and $p_o$, respectively. These can be easily experimented with, using our publicly available code\footnote[1]{\url{https://github.com/pinouche/gp_dr}}. Note that the genetic operators, maximum tree size, and tree depth are applied to single trees within a multi-tree. This means that, for instance, each tree within a multi-tree has a $p_s$ probability of having a subtree mutation. For the crossover operator, where subtrees are exchanged between two trees, one has to decide whether cross-pollination between trees at different index from the two multi-tree individuals is allowed. In~\cite{lensen2020similarity}, the authors randomly select a tree from each individual and perform standard crossover between them (random-index crossover), while in~\cite{al2021multi}, the authors only allow trees at the same position in the multi-tree to crossover with each other (same-index crossover). Here, we use same-index crossover when mixing trees from different multi-trees. The reason we make this choice is that same-index crossover encourages trees at the same index to specialize~\cite{al2021multi} and converge towards learning the same latent dimension of the lower-dimensional representation.

Finally, the population is initialized using the popular ramped half-and-half method~\cite{koza1992genetic}, and ephemeral constants are drawn from a $\mathcal{N}(0, 1)$. The terminal set (i.e., the set of all possible leaf nodes) is $\mathcal{T} = \{X, \mathcal{R}\}$, where $\mathcal{R}$ represents the set of random constants and $X = \{X^j\}_{j=1}^{p}$ is the set of variables. The function set (i.e., the set of possible non-leaf nodes) is $\mathcal{F} = \{-, +, \times\}$. This means that GP will evolve polynomials, with arbitrarily complex interactions only bounded by the depth of the tree. This results in simple expressions that can be used as a starting point to further interpretability studies on performing DR with GP. Furthermore, we conducted the same experiments using the extended function set $\mathcal{F^{'}} = \{-, +, \times, \textrm{cos}, \textrm{log}_p\}$, where $\textrm{log}(\cdot)_p=\textrm{log}(|\cdot|+\varepsilon)$, but obtained significantly worse results than when using $\mathcal{F}$. 
Due to space limitations, the results obtained with $\mathcal{F^{'}}$ are not presented here but can be found in the code repository.

\begin{table}[h!]
\centering
\caption{GP parameters and their settings.}
\label{tab:gp_parameters}

 \begin{tabular}{lc}
        \toprule  
        \textbf{Parameter}& \textbf{Value} \\
        \midrule
        Crossover rate ($p_c$)& 0.8  \\
        Subtree mutation rate ($p_s$)& 0.2  \\
        Operator muration rate ($p_o$)& 0.2\\
        Population size ($P$) & 1000 \\
        Generations & 100 \\
        Tree depth: (Min, Max)  & (2, 7)  \\
        Tournament Size & 7 \\
        \bottomrule
        \end{tabular}
\end{table}

\subsection{Mini-batch Training}

Note that during the evolutionary process, for each generation, we have to compute the distance matrix $\widetilde{D}$ for all the multi-tree individuals in the population, which has an overall computational cost of $\mathcal{O}(kn^2\times P)$, where $P$ is the population size used by GP, $n$ is the number of data points, and $k$ the number of latent variables of the lower-dimensional representation. For large values of $n$, this can be prohibitively slow. Therefore, at each generation, we randomly sample a mini-batch of $b < n$ data points from the data and compute the distance matrices on that mini-batch. In addition, mini-batch training is thought to improve generalization by changing the loss landscape at each batch, and thus escaping local optima~\cite{masters2018revisiting}.

\begin{table*}[h!]
\caption{Score of (a) balanced accuracy and (b) reconstruction error (mean $\pm$ standard deviation of 30 runs). 
For a confidence level of $1-\alpha$, statistical significance is denoted as follows: $^{*}, ^{**}, ^{***}$, for $\alpha = 0.1, 0.05, 0.01$, respectively. Statistical testing is performed between the best method and the remaining 8 methods. If the mean performance is equal across several methods, the method with the lowest standard deviation is chosen as the best method. 
The best method and those that are not statistically significantly different are in bold.
}
\centering
\label{tab:result_accuracy}
\begin{subtable}{\linewidth}
\centering
\caption{Balanced accuracy (to maximize)}
\scalebox{1.0}{
\begin{tabular}{lllllll}
\toprule  
 \multicolumn{1}{c}{} &
 \multicolumn{1}{c}{} &
 \multicolumn{5}{c}{\textbf{Datasets}} \\
\textbf{Dimensions}&\textbf{Methods}& \textbf{Segmentation} & \textbf{Ionosphere} & \textbf{Credit}  & \textbf{Libras} & \textbf{MGT} \\
\midrule
\multirow{9}{*}{$D=2$} 
& PCA &0.67$\pm$0.01 $^{***}$& 0.74$\pm$0.03 $^{***}$&\textbf{0.55$\pm$0.02} &0.28$\pm$0.04 $^{***}$&\textbf{0.66$\pm$0.01}   \\
& LLE & \textbf{0.82$\pm$0.01}&0.79$\pm$0.06 $^{*}$ &0.51$\pm$0.02 $^{***}$&\textbf{0.44$\pm$0.07} &\textbf{0.66$\pm$0.02}   \\
& Isomap &0.78$\pm$0.03 $^{***}$&\textbf{0.82$\pm$0.04}  &0.53$\pm$0.03 $^{**}$& \textbf{0.46$\pm$0.04} &0.63$\pm$0.01 $^{***}$  \\
& $\mathcal{MT},   \mathcal{F}_{\textrm{dist}}$ (Euclidean) &0.75$\pm$0.07 $^{***}$&0.78$\pm$0.04 $^{**}$ & \textbf{0.55$\pm$0.02}&0.29$\pm$0.04 $^{***}$ &\textbf{0.65$\pm$0.04}     \\
& $\mathcal{MT},   \mathcal{F}_{\textrm{dist}}$ (geodesic) &0.77$\pm$0.06 $^{*}$&0.77$\pm$0.06 $^{**}$ & 0.54$\pm$0.02 $^{*}$&0.33$\pm$0.04 $^{***}$ &0.63$\pm$0.02 $^{**}$  \\
& $\mathcal{MT},   \mathcal{F}_{\textrm{rank}}$ (Euclidean) &0.77$\pm$0.06 $^{**}$ &0.77$\pm$0.06 $^{**}$ & 0.53$\pm$0.03 $^{**}$&0.34$\pm$0.05 $^{***}$ &\textbf{0.64$\pm$0.05}   \\
& $\mathcal{MT},   \mathcal{F}_{\textrm{rank}}$ (geodesic) &0.74$\pm$0.11 $^{*}$&0.79$\pm$0.04 $^{**}$ & \textbf{0.56$\pm$0.04}&0.34$\pm$0.06 $^{***}$ &0.64$\pm$0.03$^{*}$    \\
& $\mathcal{MT},   \mathcal{F}_{\textrm{AE}}$ &\textbf{0.77$\pm$0.11} &0.79$\pm$0.05 $^{*}$ &0.51$\pm$0.02 $^{***}$& 0.35$\pm$0.04 $^{***}$&0.63$\pm$0.03 $^{**}$   \\
& $\mathcal{AMT},   \mathcal{F}_{\textrm{GP}}$ &0.77$\pm$0.07 $^{*}$&0.77$\pm$0.05 $^{**}$ &0.54$\pm$0.03 $^{*}$&0.35$\pm$0.06 $^{***}$ &0.63$\pm$0.03 $^{**}$   \\
\midrule
\multirow{9}{*}{$D=3$} 
& PCA &\textbf{0.83$\pm$0.01} & 0.83$\pm$0.03 $^{***}$&\textbf{0.57$\pm$0.03} &0.43$\pm$0.05 $^{***}$&\textbf{0.70$\pm$0.01}   \\
& LLE &\textbf{0.85$\pm$0.01} & 0.80$\pm$0.05 $^{**}$&0.51$\pm$0.02 $^{***}$&0.49$\pm$0.06 $^{*}$&\textbf{0.69$\pm$0.02}   \\
& Isomap &\textbf{0.84$\pm$0.02} & \textbf{0.84$\pm$0.03} &0.53$\pm$0.03 $^{**}$&\textbf{0.52$\pm$0.04} &\textbf{0.69$\pm$0.01}   \\
& $\mathcal{MT},   \mathcal{F}_{\textrm{dist}}$ (Euclidean) &\textbf{0.82$\pm$0.07} & 0.81$\pm$0.05 $^{**}$&\textbf{0.58$\pm$0.05} &0.45$\pm$0.05 $^{***}$&\textbf{0.71$\pm$0.04}    \\
& $\mathcal{MT},   \mathcal{F}_{\textrm{dist}}$ (geodesic) &\textbf{0.85$\pm$0.04} & 0.82$\pm$0.04 $^{**}$&\textbf{0.54$\pm$0.04} &0.42$\pm$0.06 $^{***}$&\textbf{0.69$\pm$0.05}   \\
& $\mathcal{MT},   \mathcal{F}_{\textrm{rank}}$ (Euclidean) &\textbf{0.83$\pm$0.06} & 0.83$\pm$0.04 $^{**}$&0.54$\pm$0.04 $^{*}$ &0.48$\pm$0.07 $^{**}$&\textbf{0.70$\pm$0.05}   \\
& $\mathcal{MT},   \mathcal{F}_{\textrm{rank}}$ (geodesic) &\textbf{0.85$\pm$0.05} & 0.81$\pm$0.04 $^{**}$ &0.54$\pm$0.03 $^{*}$ &0.46$\pm$0.06 $^{***}$&\textbf{0.68$\pm$0.05}    \\
& $\mathcal{MT},   \mathcal{F}_{\textrm{AE}}$ &\textbf{0.85$\pm$0.05} & 0.83$\pm$0.05 $^{*}$ &0.54$\pm$0.02 $^{*}$ &0.45$\pm$0.08 $^{***}$&\textbf{0.68$\pm$0.03}   \\
& $\mathcal{AMT},   \mathcal{F}_{\textrm{GP}}$ &\textbf{0.83$\pm$0.05} & 0.83$\pm$0.04 $^{*}$&0.53$\pm$0.04 $^{**}$ &0.46$\pm$0.04 $^{***}$&\textbf{0.69$\pm$0.04}    \\
\bottomrule
\end{tabular}
}
\end{subtable}

\vspace*{0.9cm}

\begin{subtable}{\linewidth}
\centering
\caption{Reconstruction error (to minimize)}
\scalebox{1.0}{
\begin{tabular}{lllllll}
\toprule  
 \multicolumn{1}{c}{} &
 \multicolumn{1}{c}{} &
 \multicolumn{5}{c}{\textbf{Datasets}} \\
\textbf{Dimensions}&\textbf{Methods}& \textbf{Segmentation} & \textbf{Ionosphere} & \textbf{Credit}  & \textbf{Libras} & \textbf{MGT} \\
\midrule
\multirow{9}{*}{$D=2$} 
& PCA &\textbf{0.73$\pm$0.06} &\textbf{1.27$\pm$0.13}  &0.94$\pm$0.03 $^{***}$&\textbf{0.95$\pm$0.09} &\textbf{0.63$\pm$0.02}   \\
& LLE &1.01$\pm$0.14 $^{***}$&\textbf{1.30$\pm$0.13}  &0.96$\pm$0.03 $^{***}$&1.01$\pm$0.09 $^{*}$&0.81$\pm$0.06 $^{***}$  \\
& Isomap &0.81$\pm$0.10 $^{**}$&\textbf{1.26$\pm$0.14} &\textbf{0.88$\pm$0.03} &\textbf{0.97$\pm$0.10} &\textbf{0.63$\pm$0.03}   \\
& $\mathcal{MT}, \mathcal{F}_{\textrm{dist}}$ (Euclidean) &0.79$\pm$0.12 $^{*}$&\textbf{1.29$\pm$0.13}  &0.96$\pm$0.04 $^{***}$&\textbf{0.97$\pm$0.11} &0.66$\pm$0.04 $^{*}$  \\
& $\mathcal{MT}, \mathcal{F}_{\textrm{dist}}$ (geodesic) &0.79$\pm$0.12 $^{*}$&\textbf{1.30$\pm$0.13}  &0.94$\pm$0.04 $^{***}$&\textbf{0.98$\pm$0.10} &0.68$\pm$0.04  $^{***}$\\
& $\mathcal{MT}, \mathcal{F}_{\textrm{rank}}$ (Euclidean) &\textbf{0.79$\pm$0.15} &1.32$\pm$0.14 $^{*}$ &0.96$\pm$0.03 $^{***}$&\textbf{0.96$\pm$0.10} &0.69$\pm$0.05 $^{***}$  \\
& $\mathcal{MT}, \mathcal{F}_{\textrm{rank}}$ (geodesic) &0.86$\pm$0.19 $^{**}$&1.33$\pm$0.15 $^{**}$ &0.97$\pm$0.05 $^{***}$&\textbf{0.96$\pm$0.10} &0.72$\pm$0.05 $^{***}$  \\
& $\mathcal{MT}, \mathcal{F}_{\textrm{AE}}$ &\textbf{0.72$\pm$0.08} &\textbf{1.29$\pm$0.14} &\textbf{0.89$\pm$0.03} &0.95$\pm$0.09 &0.66$\pm$0.04 $^{**}$  \\
& $\mathcal{AMT}, \mathcal{F}_{\textrm{GP}}$ &0.78$\pm$0.09 $^{*}$&\textbf{1.28$\pm$0.13}  &0.95$\pm$0.04 $^{***}$&\textbf{0.94$\pm$0.09} &\textbf{0.63$\pm$0.03}   \\
\midrule
\multirow{9}{*}{$D=3$} 
& PCA &\textbf{0.60$\pm$0.08} &\textbf{1.24$\pm$0.14}  &0.90$\pm$0.04 $^{***}$&\textbf{0.89$\pm$0.10} &0.62$\pm$0.02 $^{*}$  \\
& LLE &0.98$\pm$0.15 $^{***}$& \textbf{1.28$\pm$0.14} &0.94$\pm$0.03 $^{***}$&0.95$\pm$0.10 $^{**}$&0.77$\pm$0.04 $^{***}$  \\
& Isomap &0.74$\pm$0.14 $^{**}$& \textbf{1.25$\pm$0.14} &\textbf{0.85$\pm$0.03} &0.97$\pm$0.08 $^{**}$&\textbf{0.62$\pm$0.03}   \\
& $\mathcal{MT}, \mathcal{F}_{\textrm{dist}}$ (Euclidean) &0.69$\pm$0.09 $^{**}$&\textbf{1.28$\pm$0.14}  &0.92$\pm$0.04 $^{***}$&\textbf{0.89$\pm$0.11} &0.64$\pm$0.02 $^{***}$    \\
& $\mathcal{MT}, \mathcal{F}_{\textrm{dist}}$ (geodesic) &0.75$\pm$0.13 $^{***}$&\textbf{1.30$\pm$0.15}  &0.92$\pm$0.04 $^{***}$&\textbf{0.92$\pm$0.10} &0.65$\pm$0.04 $^{**}$  \\
& $\mathcal{MT}, \mathcal{F}_{\textrm{rank}}$ (Euclidean) &0.73$\pm$0.11 $^{***}$&1.31$\pm$0.13 $^{*}$ &0.94$\pm$0.06 $^{***}$&\textbf{0.89$\pm$0.10} &0.65$\pm$0.04 $^{***}$  \\
& $\mathcal{MT}, \mathcal{F}_{\textrm{rank}}$ (geodesic) &0.74$\pm$0.16 $^{**}$&1.34$\pm$0.17 $^{*}$ &0.93$\pm$0.08 $^{**}$&\textbf{0.87$\pm$0.07} &0.69$\pm$0.04 $^{***}$   \\
& $\mathcal{MT}, \mathcal{F}_{\textrm{AE}}$ &0.67$\pm$0.08 $^{**}$&\textbf{1.26$\pm$0.14} &0.88$\pm$0.03 $^{*}$&\textbf{0.86$\pm$0.09} &0.63$\pm$0.02 $^{***}$   \\
& $\mathcal{AMT}, \mathcal{F}_{\textrm{GP}}$ &0.72$\pm$0.11 $^{**}$& \textbf{1.26$\pm$0.14} &0.91$\pm$0.03 $^{***}$&\textbf{0.86$\pm$0.09}&\textbf{0.60$\pm$0.03}    \\
\bottomrule
\end{tabular}
}
\end{subtable}
\end{table*}

\subsection{Evaluation Protocol}
\label{sec:eval_protocol}

A detailed account of the overall evaluation process, from the original data to the final predictions, is given in Figure \ref{fig:evaluation_protocol}. First, the original data is split into two subsets: (i) the training data for the DR model $M$, and, (ii) a held-out set to apply $M$ on and obtain its lower-dimensional representation. In turn, this lower-dimensional data is split in two subsets: (i) the training data for the evaluation model $E$, and, (ii) the hold-out set to apply the evaluation model $E$ to compute the performance metrics. The evaluation model $E$ (e.g., a neural decoder for reconstruction error and a random forest classifier for accuracy) is trained using a 10-fold cross validation and we report the results on the test set (see Section \ref{sec:results}). Note that the models used for the evaluation phase are described in Section \ref{sec:eval_functions}.

\section{Results}
\label{sec:results}

\subsection{Quantitative Analysis}

In Table \ref{tab:result_accuracy} (a), we display the average balanced accuracy and associated standard deviation computed across 30 independent runs for all the GP-based methods and the baselines. Similarly, in Table \ref{tab:result_accuracy} (b), we display the results for the reconstruction error. In addition, we compute statistical significance using the Mann-Whitney U rank test~\cite{mann1947test}, denoted by the asterisk notation.

We can see that the non-linear baseline methods LLE and isomap seem to yield the better performances, in particular when $k=2$. For $k=3$, $\mathcal{F}_{\textrm{dist}}$ (Euclidean) performs best on 2 of the datasets but is not always statistically significant. Overall, it seems that for $k=3$, the difference between the methods is less pronounced than for $k=2$. This is intuitive since the more dimensions we keep the easier it is to capture relevant information from the original data, thus attenuating potential differences between methods. In fact, for more complex datasets with a higher number of features, such as Ionosphere and Libras Movement, we can still see a significant difference between the best performing method (isomap) and the other methods. Additionally, PCA (the only linear dimensionality reduction method in our experiments) does not perform well on these two datasets, suggesting that non-linearity allows for capturing relevant relationships between the original variables. 

\begin{table*}[t]
\centering
\caption{Expressions of the lower-dimensional representation of the best solution in terms of reconstruction error (out of the 30 independent runs), for all GP-based methods on the Libras and Segmentation datasets
from UCI~\cite{Dua:2019}. 
}
\label{tab:expressions}
\scalebox{0.83}{
\begin{tabular}{llcc}
\toprule  
 \multicolumn{1}{c}{} &
  \multicolumn{1}{c}{} &

 \multicolumn{2}{c}{\textbf{Datasets}} \\
\textbf{Method}& \textbf{Dim} & \textbf{Segmentation} & \textbf{Libras Movement} \\
\midrule

\multirow{2}{*}{$\mathcal{F}_{\textrm{dist}}$ (Euclidean)}
& $\widetilde{X}^1$ & $x_{18} - x_{6} - x_{8}$ & $-x_{14} - x_{26} + x_{45} + x_{55}$ \\
& $\widetilde{X}^2$ & $x_{14} - x_{17} - x_{2}$ & $x_{55}+x_{58}$ \\
\midrule
\multirow{2}{*}{$\mathcal{F}_{\textrm{dist}}$ (geodesic)}
& $\widetilde{X}^1$ & $x_{0} + 3x_{14} + x_{7}$ & $-x_{11}x_{67} + x_{42} + x_{62} + x_{70}$   \\
& $\widetilde{X}^2$ & $x_{15}x_{18} + x_{15} + x_{8}$ & $x_{0}^{2}x_{36}$ \\
\midrule
\multirow{2}{*}{$\mathcal{F}_{\textrm{rank}}$ (Euclidean)}
& $\widetilde{X}^1$ & $x_{18}$ & $x_{11} + x_{15} + x_{16} + x_{42} + x_{52}$ \\
& $\widetilde{X}^2$ & $x_{14} - x_{17}(x_{5} + x_{9}) - x_{18} + x_{2}$ & \hspace{0.3cm}$x_{20}(x_{18} + x_{53} - x_{55} + x_{7} - x_{74}) - x_{85} + 0.561$ \\
\midrule
\multirow{2}{*}{$\mathcal{F}_{\textrm{rank}}$ (geodesic)}
& $\widetilde{X}^1$ & $x_{11} - x_{15} - x_{16} - x_{6}$ & $x_{10}x_{78} + x_{12} - x_{80}$ \\
& $\widetilde{X}^2$ & $0.767x_{14} - x_{18}$ & $x_{19} + x_{20} + x_{42} + x_{72} - x_{74}$ \\
\midrule
\multirow{2}{*}{$\mathcal{F}_{\textrm{AE}}$}
& $\widetilde{X}^1$ & $-x_{10} - x_{12} + x_{18} - x_{6}$ & $x_{0} - x_{20}x_{86} + x_{7}$  \\
& $\widetilde{X}^2$ & $x_{12}x_{5}$ &  $x_{16}x_{53} + x_{29} + x_{59}$  \\
\midrule
\multirow{2}{*}{$\mathcal{F}_{\textrm{GP}}$}
& $\widetilde{X}^1$ & $x_{11} + x_{8}$ & $-x_{28} + x_{40} + x_{53} + x_{86}$\\
& $\widetilde{X}^2$ & $x_{2}^{2} + x_{6}$ & $-x_{15} + x_{20} + x_{59} + x_{73}$ \\
\bottomrule
\end{tabular}
}
\end{table*}

In Table \ref{tab:result_accuracy} (b), we can see that $\mathcal{F}_{\textrm{AE}}$ and $\mathcal{F}_{\textrm{GP}}$ have the lowest reconstruction error out of the GP-based methods. This is expected because these methods minimize the reconstruction error during the optimization process. In addition, we can see that PCA and isomap yield good results across several datasets. Interestingly, while LLE performs well when balanced accuracy is used as metric, it is arguably the worst performing method to reconstruct the original data. This may be explained by the fact that co-linear features are equally weighted when computing the distance matrices $D$ and $\widetilde{D}$. This is detrimental to maximizing accuracy as there may be independent variables (e.g., non co-linear) that are important for the downstream classification task but which would be out-weighted when computing the distance matrices. For this reason, we performed PCA before computing distances, as mentioned in Section \ref{sec:pca_init}. However, we only experimented with keeping $99\%$ of the original variance and one would need to experiment further to identify the effect of performing PCA on the balanced accuracy and reconstruction error. Finally, while $\mathcal{F}_{\textrm{AE}}$ and $\mathcal{F}_{\textrm{GP}}$ do perform well, there exists drawbacks to these methods. For $\mathcal{F}_{\textrm{AE}}$ we have to train a neural network autoencoder, which adds a layer of complexity to the overall method. That is, if our autoencoder teacher does not perform well in the first place, the GP-based student is bound to fail. On the other hand, while $\mathcal{F}_{\textrm{GP}}$ does not require an autoencoder teacher model, it does not scale well with the number of input features since the GP-decoder requires a number of trees equal to the number of input features, as depicted in Figure \ref{fig:tree_representation}c.

Overall, our results show that GP-based dimensionality reduction methods can be on a par with proven baseline methods. 
In addition, a major advantage of GP-based methods is that they have the potential to produce interpretable mappings in the form of symbolic expressions, which we discuss next.

\subsection{Qualitative Analysis}
\label{sec:expressions}

In Table \ref{tab:expressions}, we display the expressions that map the original data into a lower-dimensional representation, for the various GP-based methods. We can see that the resulting embeddings are small enough to have the potential of being interpretable. This is in contrast to the baseline methods where only PCA provide a global functional mapping from the original data to its lower-dimensional representation. However, PCA strictly returns a linear combination of all the variables, limiting its complexity and rendering it unintelligible when the number of input variables is large. On the other hand, GP-based methods can construct non-linear expressions while only using a small subset of the original variables. 

\section{Discussion}
\label{sec:conclusion}

In this paper, we have seen that GP can be competitive with widely used DR methods, while producing small, non-linear, and potentially interpretable functional mappings of how the original data can be compressed into the latent dimensions. 
In particular, when the fitness used for GP is the same as the objective we want to optimize for, (e.g., reconstruction error), 
GP often outperforms baselines. 

A limitation of this paper is that 
we adopted classic 
GP
evolutionary operators for variation and selection, yet  
state-of-the-art GP methods 
include more interesting mechanisms.
For instance, one could use a GP version that is more expressive such as differentiable Cartesian GP~\cite{izzo2017differentiable} (DCGP). This is because in DCGP, one can optimize internal weights of the expression using gradient descent. 
Another example could be to adopt GP-GOMEA~\cite{virgolin2021improving} as it is state-of-the-art to discover accurate yet small expressions~\cite{la2021contemporary}.

Another limitation of this study is that we only used a fixed set of parameters and did not study how different GP representations and fitness functions inter-relate to different evolutionary budgets and settings.
For example, we used a classic population size of 1000.
However, the population size is a key parameter because it determines the supply of (high-order) building blocks available for the evolution. 
Some recent works in GP such as~\cite{schweim2021sampling,virgolin2021improving} suggest that better performance is only achievable when the population size is of tens of thousands or more.
In light of this, future work on understanding the potential of GP for DR should include an analysis of the effect of important parameter settings.
Moreover, future work should also include more baselines and loss functions in the comparison, including, e.g., the recenly introduced PaCMAP~\cite{wang2021understanding}.

An interesting aspect emerging from our results is that DR methods achieving high accuracy often performed worse in terms of reconstruction error (and vice versa). This can happen because, when performing reconstruction, each reconstructed feature is equally important; whereas to answer a predictive task as per classification, only a subset of them may be. 

\section{Conclusion}

In conclusion, we have considered different versions, i.e., representations and fitness functions of which two are novel (GP autoencoder representation and autoencoder teacher fitness), of genetic programming for dimensionality reduction. 
We have found that genetic programming is a competitive approach to obtain interpretable mappings for dimensionality reduction.
We thus believe that genetic programming for dimensionality reduction is a promising research avenue.


\clearpage
\vspace{5cm}

\bibliographystyle{ACM-Reference-Format.bst}
\bibliography{acmart.bib}

\end{document}